\documentclass[10pt,twocolumn,letterpaper]{article}

\usepackage[pagenumbers]{cvpr} 

\definecolor{cvprblue}{rgb}{0.21,0.49,0.74}
\usepackage[pagebackref,breaklinks,colorlinks,allcolors=cvprblue]{hyperref}

\usepackage{amsmath}
\usepackage{amssymb} 
\usepackage{booktabs}
\usepackage{adjustbox}
\usepackage{multirow}
\usepackage{color} 
\usepackage{colortbl} 

\def\paperID{*****} 
\def\confYear{2026}

\title{Leveraging Causal Reasoning Method for Explaining Medical Image Segmentation Models}

\author{
	Limai Jiang$^{1,2}$ \and
	Ruitao Xie$^{1,3}$ \and
	Bokai Yang$^{1,4}$ \and
	Huazhen Huang$^{1}$ \and
	Juan He$^{1,5}$ \and
	Yufu Huo$^{1,2}$ \and
	Zikai Wang$^{1,2}$ \and
	Yang Wei$^{1,6}$ \and
	Yunpeng Cai$^{1,*}$
	\medskip \\
	\normalsize
	$^{1}$Shenzhen Institutes of Advanced Technology, Chinese Academy of Sciences \\
		\normalsize
	$^{2}$University of Chinese Academy of Sciences \\
		\normalsize
	$^{3}$Shenzhen University of Advanced Technology \\
		\normalsize
	$^{4}$Shenzhen Polytechnic University \\
		\normalsize
	$^{5}$University of Macau \\
		\normalsize
	$^{6}$Southern University of Science and Technology \\
	\small
	{\tt lm.jiang2@siat.ac.cn}, {\tt yp.cai@siat.ac.cn}
}

\begin{document}
\maketitle

\begin{abstract}
	Medical image segmentation plays a vital role in clinical decision-making, enabling precise localization of lesions and guiding interventions. Despite significant advances in segmentation accuracy, the black-box nature of most deep models has raised growing concerns about their trustworthiness in high-stakes medical scenarios. Current explanation techniques have primarily focused on classification tasks, leaving the segmentation domain relatively underexplored. We introduced an explanation model for segmentation task which employs the causal inference framework and backpropagates the average treatment effect (ATE) into a quantification metric to determine the influence of input regions, as well as network components, on target segmentation areas. Through comparison with recent segmentation explainability techniques on two representative medical imaging datasets, we demonstrated that our approach provides more faithful explanations than existing approaches. Furthermore, we carried out a systematic causal analysis of multiple foundational segmentation models using our method, which reveals significant heterogeneity in perceptual strategies across different models, and even between different inputs for the same model. Suggesting the potential of our method to provide notable insights for optimizing segmentation models. Our code can be found at \url{https://github.com/lcmmai/PdCR}.
\end{abstract}

\section{Introduction}

A vast number of deep learning-based methods have been proposed in the field of medical imaging, integrating various meticulously designed modules to address the complexity inherent in diverse clinical scenarios. While these methods often demonstrate measurable improvements in experimental metrics, it remains unclear what features they truly capture, whether they behave as intended, and what underlying factors drive their predictions. Consequently, these models are perceived as ``black boxes"~\cite{b0015}.

With the continuous development of artificial intelligence (AI), merely focusing on model outputs is no longer sufficient to meet the needs of users and clinicians. An increasing number of studies aim to reveal the internal decision-making processes of models, in order to achieve transparent and explainable AI (XAI)~\cite{b0016}. Image classification and segmentation are the cornerstones of clinical applications, and their interpretability is crucial for downstream diagnosis and decision-making. Among them, classification tasks have seen significant progress in related interpretability methods~\cite{b003,b007}. However, segmentation is a dense prediction task, where one input variable may contribute to multiple aspects of a complex high dimensional output structure as well as the corresponding network activations, and an output variable may also depend on the value of other outputs. As a result, existing methods are difficult to extend to segmentation models that involve both complex spatial reasoning and fine-grained decisions, leading to a serious lack of systematic tools and methods at present~\cite{b0011}.

Despite several prior attempts to adapt traditional XAI methods for segmentation, prior research has predominantly focused on final performance metrics or correlation analysis~\cite{b010}, often overlooking the underlying mechanisms and causal relationships. These methods, including permutation-based and gradient-based ones, attempt to identify the most relevant regions and implicitly assume these areas bear full causal responsibility for the outcomes~\cite{b009}. But it is well known that correlation does not necessarily imply causation. Especially for segmentation problems, the large number of input-output interactions and the complex latent interdependent structure within both input and output features would create many false connections, leading to inaccurate or inefficient explanations. Moreover, In light of the slowing growth in computational resources~\cite{b008}, it becomes increasingly important to re-examine the actual roles of network modules, rather than relying solely on scaling and data volume to improve performance.

This gap calls for an explainable approach capable of revealing the causal mechanisms linking model inputs and outputs. To address this, we propose \textbf{P}erturbation-\textbf{d}riven \textbf{C}ausal \textbf{R}easoning (PdCR). Specifically, given a trained black-box model and an arbitrary input, we first identify a region of interest (RoI). By deliberately introducing perturbations to induce contextual bias around this region, we then quantify its effect on the output to infer causal influence, whether positive or negative. Through visualizing these effects, PdCR enables direct observation of the model’s true functional behavior, offering deeper insights. Our contributions can be summarized as follows:

\begin{itemize}
	
	\item We propose PdCR, a novel model-agnostic framework designed to explain segmentation models by quantifying the causal influence of input regions on predictions. Unlike current perturbation-based approaches, we execute a collection of concurrent perturbations and attribute the collective responses back to each features.  
	
	\item By adopting the idea of average treatment effect (ATE) in causal inference, our method systematically measures how perturbations surrounding RoI alter the segmentation output, enabling causality-driven bidirectional attribution that reveals both positively and negatively contributing areas.
	
	\item Extensive experiments are conducted on two distinct types of medical imaging datasets, applying a variety of representative methods. PdCR achieves advanced interpretability performance compared with existing XAI approaches. Moreover, it also helps with revealing the distinctions in perception patterns across different networks.

\end{itemize}

As one of the few explainability methods tailored for segmentation models, we hope that the proposed PdCR can serve as a solid foundation in this area, offering new analytical perspectives for various tasks, assisting researchers in validating the practical effectiveness of different mechanisms, and deepening the understanding of medical image segmentation networks.

\section{Related Work}

\subsection{Medical Image Segmentation}

Over the years, segmentation models have evolved from early convolutional neural networks (CNNs)~\cite{b015}, to more complex architectures. For example, CNN-based models exploit local patterns~\cite{b051,b052,b053}; ViT-based models capture global dependencies~\cite{b054,b019,b020}; newer designs (e.g., MLPs~\cite{b055}, Mamba~\cite{b022,b023,b026}, KAN~\cite{b021}) explore efficient sequential or nonlinear dynamics. Despite their architectural diversity, these methods typically follow a bottom-up pipeline, mapping pixel-level features to structured output masks.

\subsection{Explainability in Medical Image}

In high-stakes domains like healthcare, model explainability is crucial for trust and safe clinical adoption. Most XAI methods in medical imaging focus on classification tasks. Post-hoc techniques like class activation mapping~\cite{b028,b005} highlight image regions relevant to diagnostic decisions. Some methods incorporate language models to mimic physician reasoning~\cite{b003,b030}, while counterfactual approaches reveal decision boundaries by modeling data distributions~\cite{b031}.

In contrast, medical image segmentation involves spatially detailed outputs, where each pixel contributes to a complex structured prediction~\cite{b0092}. Explaining such outputs requires identifying influential regions and understanding how global context shapes each RoI. Some XAI methods embed slightly transparent~\cite{b0013,b0014} or visualizable~\cite{b0012} modules into segmentation models to build interpretable decision processes. Some methods attempt post-hoc explanations, while they often focus on minimal highlighting segmented areas~\cite{b009} or use correlation-based attributions~\cite{b0093}, offering limited insight into the model’s reasoning. A causally approach to interpret segmentation models remains largely absent.
﻿
﻿

\begin{figure*}[t]
	\centering
	\includegraphics[width=1\linewidth]{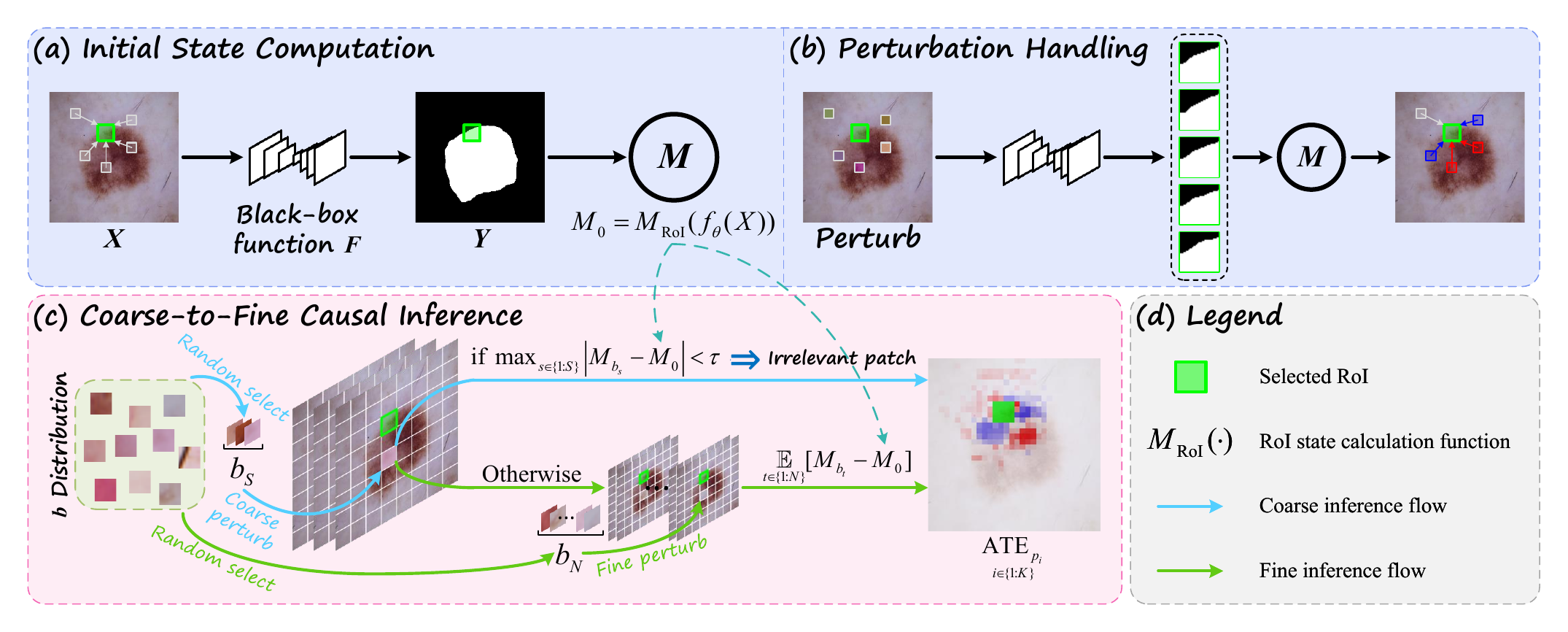}
	\caption{The overall workflow of PdCR. (a) Given an image $X$, an RoI is selected and its initial state $M_0$ is computed. The surrounding region (gray box) may influence it. (b) Perturbations are applied to $X$, and outputs are compared to assess the causal impact. (c) Patch-wise perturbations are performed in a coarse-to-fine manner. A small set $b_N$ is first used to filter out irrelevant patches, followed by a larger set $b_N$ to compute a fine-grained causal saliency map. (d) Legend for symbols and lines.}
	\label{fig_network}
\end{figure*}

\subsection{Causal Reasoning}

Causality reflects the intrinsic properties and structural relationships within and among variables, and causal reasoning aims to uncover these underlying connections~\cite{b035}. Moreover, due to its mechanistic advantages, it has been introduced into the field of computer vision and has achieved progress in various tasks such as causal inquiry~\cite{b041}, visual generation~\cite{b040}, medical visual question answering~\cite{b044}, and image classification~\cite{b0094}. However, its application in medical image segmentation remains in its infancy~\cite{b0095}. Integrating causal reasoning into segmentation tasks to enhance generalization and interpretability is a promising and worthwhile direction for future research.

\section{Method}

PdCR is designed to quantify the causal impact of different regions on RoI’s result from black-box models. As illustrated in Fig.~\ref{fig_network}, it proceeds through four stages: (1) choosing an RoI and measuring its initial state; (2) selecting a series of blocks and carry out perturbations on surrounding patches; (3) evaluating the causal effects of each patch; (4) iteratively filter and refine patches by pruning irrelevant regions, and repeat (2)-(4) on modified patches to improve efficiency. Although perturbation is already adopted in existing XAI methods, our method is the first one to process it in a causal inference framework specifically for segmentation models, bringing entirely different explanation principles and more precise feature weightings. Here we go on to explain the motivation behind this design and details.

\subsection{Preliminaries}

The goal of causal reasoning is to evaluate the effect of one causal factor on another variable~\cite{b045}. In visual tasks, the process can be considered as a causal structure $X \xrightarrow{F} Y \to M$. An image $X$ is processed through a complex black-box mapping function $F$, resulting in an output $Y$, which may represent class labels, segmentation maps, or other task-specific predictions. The quality of the output is evaluated by a performance metric $M$ (e.g., similarity).

We model interventions on images within the framework of the potential outcome model~\cite{b047}. Ideally, the difference between potential outcomes can be interpreted as the causal effect of a treatment on the result. For instance, given a binary treatment $T=0/1$, each individual subject $I$ can receive only one treatment in practice, denoted by the causal intervention $do(I=0/1)$~\cite{b046}, leading to corresponding outcomes $R_{0/1}$. The individual treatment effect (ITE) is then defined as $R_1 - R_0$. Due to individual variability or other confounding factors, this estimate may be biased. To mitigate such effects, the analysis can be extended to a population of size $N$, allowing for the computation of the ATE as a more robust measure:
\begin{equation}
	\label{eq01}
	\text{ATE} = \mathbb{E}[R_1 - R_0] = \frac{1}{N} \sum_{i=1}^{N} \left(R_1 - R_0 \right).
\end{equation}
Here, $I$ corresponds to the image $X$, and $R$ corresponds to the evaluation metric $M$ in the visual task formulation.

\subsection{Intervening Black-box Models}

Based on the above motivation and under a model-agnostic setting, we focus on observable variables $X$, $Y$, and $M$. Our goal is to reveal the causal effect of $X$ on $M$ through controlled interventions using the $do(\cdot)$ operator. This process follows the causal path: $do(X) \xrightarrow{F} Y \rightarrow M$.

Concretely, PdCR analyzes variations through perturbation-based destruction. Given a function $F$, i.e., an arbitrarily parameterized network $f_\theta$ with parameters $\theta$ and a medical image as input, an RoI is selected as the observation target, while the remaining regions are perturbed. Notably, perturbing the entire remaining image globally disrupts all contextual structures, leading to abnormal model behavior, which is both unrealistic and uninformative. Motivated by the most natural intuition, we partition the image into a set of small patches ${X_{p_i}}, i \in \{ 1:K\}$ and perform interventions on a single patch at a time, independently of the others. The next step is to identify minimal yet effective perturbations $do(\cdot)$ that yield meaningful insight.

\begin{figure}
	\centering
	\includegraphics[width=1\linewidth]{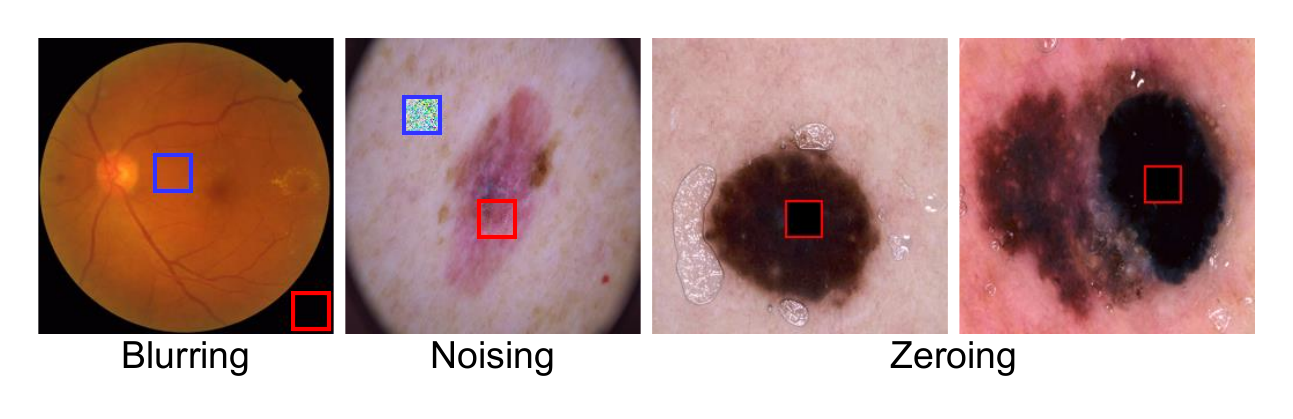}
	\caption{Examples of original image and perturbations applied to its patches.}
	\label{oriperturb}
\end{figure}

\subsection{Principles of Intervention}

We first establish the criteria for selecting valid perturbations: (1) \textbf{Contextual Disruption}. Contextual relationships are inherently present in real-world data; therefore, a valid perturbation must disrupt these relationships and lead to substantive changes in the content of the affected patch. (2) \textbf{Distributional Plausibility}. The perturbed image should remain within the distribution of the test sample space, avoiding damage that pushes the image into an out-of-distribution regime. (3) \textbf{Attributability}. Perturbations should have a clear target and produce traceable effects. The analysis is only meaningful when the model's response can be reliably attributed to the localized intervention.

Previous methods have employed various perturbations such as blurring, noise addition, and zeroing out~\cite{b048}. However, these perturbations may fail in the current context. We illustrate some potential shortcomings of these perturbations in Fig.~\ref{oriperturb}. In medical images, intra-patient regions often lack rich textures, and dark corners or black pixels are much more prevalent compared to natural images. For example, applying Gaussian blur within the blue box in the first image results in negligible change, and blurring or zeroing the background in the red box at the bottom right is evidently meaningless. Nevertheless, empirically, some models might rely on such background cues for localization, which calls for meaningful perturbations. In the second image, noise is added separately within the blue and red boxes; noise at too small a scale fails to induce any change, while excessive noise damages the image. Moreover, certain lesions are inherently nearly pure black or monochromatic, rendering these perturbations even less effective. Although synthetic toy examples~\cite{b049} might be feasible, measuring the correlation of such biases in real-world practice remains challenging.

To mitigate limitations, we opt to extract perturbations from the natural distribution corresponding to the dataset. Specifically, when perturbing a dermatological image, we first partition a subset from the dataset that excludes the target image. The intervention $do(\cdot)$ is then defined by randomly cropping and extracting a block $b$ from this subset to replace the original region. In the segmentation setting, the output $Y$ must be evaluated by the model $M$. As illustrated in Fig.~\ref{fig_network}(a) and (b), let the model’s response to the initial RoI be denoted as $M_0={M_{RoI}}({f_\theta }(X))$. The perturbation applied to the $i$-th patch is represented as $\left. X \right|do({X_{p_i}}=b)$. ${\rm{ITE}}{p_i}$ can be computed and substituted into Eq.~\ref{eq01}. By performing $N$ such interventions, ${\rm{ATE}}{p_i}$ is computed as:
\begin{equation}
	\label{eq02}
	\resizebox{0.45\textwidth}{!}{$
		\begin{aligned}
			\mathop {{\rm{AT}}{{\rm{E}}_{{p_i}}}}\limits_{i \in \{ 1:K\}} 
			&= \mathop{\mathbb{E}} \limits_{t \in \{ 1:N\} } [M_{b_t} - M_0] \\
			&= \frac{1}{N} \sum_{t = 1}^N \left( M_{\mathrm{RoI}}\left(f_\theta\left(X \mid do(X_{p_i} = b_t)\right)\right) - M_{\mathrm{RoI}}(f_\theta(X)) \right)
		\end{aligned}
		$},
\end{equation}
where $K$ denotes the number of patches the image is divided into, $N$ is a user-defined parameter, and $M$ can be instantiated with any commonly used segmentation metric. In this work, we adopt the standard Dice similarity coefficient (DSC)~\cite{b050}.

\subsection{Pruning Strategy}

Calculating the ATE for a single patch may not take much time; however, the computational scale and the number of image patches increase exponentially as the patch size decreases. For example, the input size of medical image models is typically $256 \times 256$. Dividing it into $8 \times 8$ patches results in 1024 patches, while dividing it into $4 \times 4$ patches produces 4096 patches. Assuming $N=200$, this leads to 819,200 computations. Therefore, it is necessary to consider pruning the computation tree to reduce the time cost.

\subsubsection{Coarse-to-Fine Patch Screening}

The causal effect of the RoI on itself is infinite, but it is not sensitive to perturbations at every other location. In particular, changes in pixels that are far away may have little influence on the model’s judgment of the RoI. Therefore, interventions can be performed in a coarse-to-fine manner. Specifically, a patch is first substituted $S$ times (set to $3$ in this work). If all the resulting ITE values are below the threshold $\tau$ (set to $0.02$ in this work), the patch is considered irrelevant to the RoI and thus ignored. According to Fig.~\ref{fig_network}(c), all patches deemed relevant will proceed to the full ATE calculation:
\begin{equation}
	\label{eq03}
	\resizebox{0.45\textwidth}{!}{$
		\mathop {{\rm{AT}}{{\rm{E}}_{{p_i}}}}\limits_{i \in \{ 1:K\}}  = \left\{ {\begin{array}{*{20}{l}}
				{ + \infty ,}&{{\rm{if\enspace}}{X_{{p_i}}} \in {\rm{RoI}}}\\
				{0,}&{{\rm{if\enspace}}{{{\max }_{s \in \{ 1:S\} }}} \left| {{M_{{b_s}}} - {M_0}} \right| < \tau }\\
				{\mathop{\mathbb{E}} \limits_{t \in \{ 1:N\} } [{M_{{b_t}}} - {M_0}],}&\text{otherwise}
		\end{array}} \right.
		$}.
\end{equation}

\subsubsection{Convergence Steps}

\begin{figure}
	\centering
	\includegraphics[width=1\linewidth]{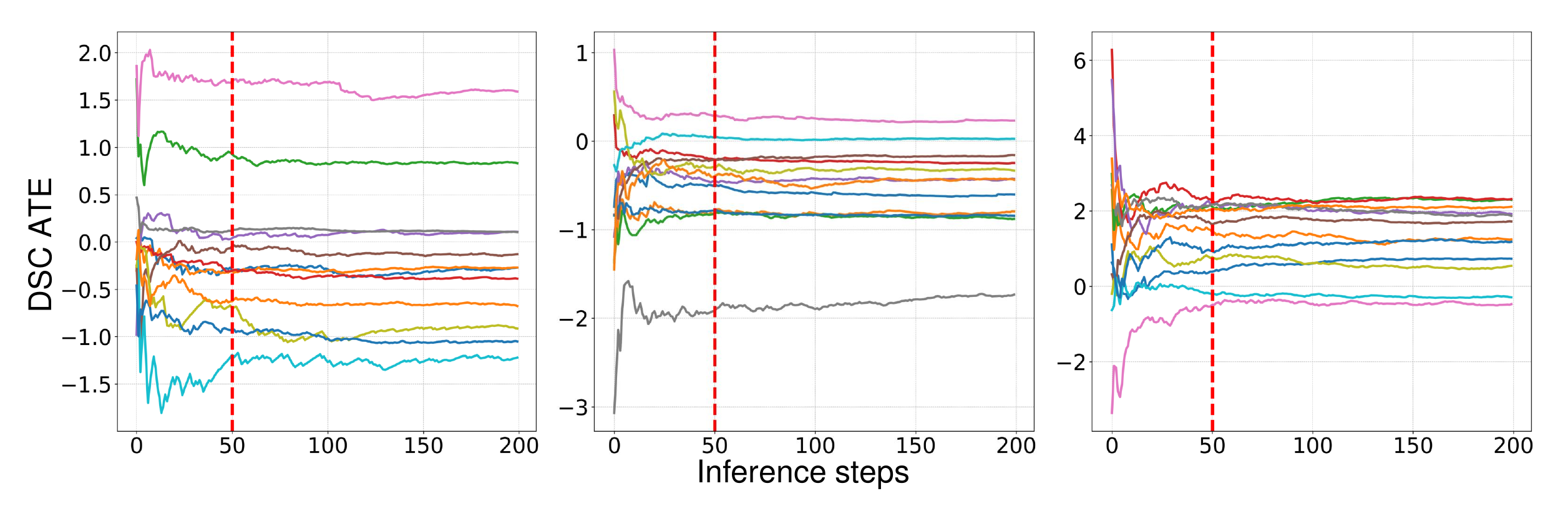}
	\caption{Relationship between the number of inference steps and the results, where lines of different colors represent different networks.}
	\label{infersteps}
\end{figure}

A highly confident causal relationship may require many inference steps, which inevitably prolongs the overall procedure. Therefore, it is necessary to determine a value of $N$ beyond which further increases yield diminishing returns and do not introduce significant bias. Ultimately, $N=50$ was chosen, as shown in Fig.~\ref{infersteps}. The inference results initially fluctuate dramatically but gradually stabilize as the number of steps increases. It can be observed that around step 50, the DSC ATE for the majority of networks converges, indicating that additional steps beyond this point do not lead to substantial differences.


In summary, a coarse-to-fine estimation is conducted to assess the impact of all image patches on RoI segmentation performance, based on these results, we construct a PdCR map. Regions exhibiting negligible influence are assigned a zero value. If perturbing a region leads to a degradation in RoI segmentation performance, the region is considered to have a positive contribution and is marked in red. Conversely, if performance improves after perturbation, the region is assigned a negative contribution and displayed in blue. The color intensity reflects the magnitude of the contribution, with deeper colors indicating stronger causal effects.

\section{Experiments}

This section outlines the experimental setup and compares PdCR with other segmentation explainability methods to demonstrate its advantages. We then use PdCR to analyze model behaviors and discuss underlying mechanisms.

\subsection{Experimental Configuration}

\subsubsection{Black-box Segmentation Models and Baseline Explanation Methods}

To illustrate the evolution of medical image segmentation, we analyze 12 representative methods spanning eleven years, as listed in Table~\ref{segmentation_methods} with their publication years and underlying mechanisms. These include CNNs, MLPs, ViTs, Mamba, KANs, and hybrid architectures. Meanwhile, two interpretability methods, SEG-GRAD~\cite{b0093} and MiSuRe~\cite{b009}, are introduced to compare the performance of PdCR.

\begin{table}[t!]
	\centering
	\small
			\begin{tabular}{l|cc}
				\toprule
				\rowcolor[gray]{0.9}
				\textbf{Method} & \textbf{Year} & \textbf{Mechanism} \\
				\midrule
				U-Net~\cite{b015}           & 2015          & CNN                \\
				SegNet~\cite{b051}          & 2017          & CNN                \\
				UNet++~\cite{b052}          & 2018          & CNN                \\
				HRNet~\cite{b053}           & 2020          & CNN                \\
				SegFormer~\cite{b054}       & 2021          & ViT                \\
				TransUNet~\cite{b020}       & 2021          & ViT + CNN          \\
				Swin-Unet~\cite{b019}       & 2022          & ViT (Swin)         \\
				UNeXt~\cite{b055}           & 2022          & MLP + CNN          \\
				VM-UNet~\cite{b022}         & 2024          & Mamba              \\
				LightM-UNet~\cite{b023}     & 2024          & Mamba              \\
				U-KAN~\cite{b021}           & 2025          & KAN + CNN          \\
				MCU-RE~\cite{b026}          & 2025          & Mamba + CNN        \\
				\bottomrule
			\end{tabular}
			\caption{Overview of representative semantic segmentation methods, including their publication years and architectural mechanisms.}
			\label{segmentation_methods}
		\end{table}

\subsubsection{Datasets and Setting Details}

For both model training and causal inference, two representative datasets are used: (1) \textbf{HAM10000}~\cite{b056}: skin lesion segmentation. Lesions are relatively large, continuous, diverse, and irregular in shape, with blurred boundaries. The dataset includes 8,000 training images, 1,965 for cropping perturbation blocks, and an additional 50 for causal reasoning. (2) \textbf{FIVES}~\cite{b057}: retinal vessel segmentation. Vessels are extremely thin, tree-like, widely distributed, and often contain discontinuous capillaries. The dataset contains 600 training images, 170 for perturbation block cropping, and 28 for causal inference.


\begin{figure*}
	\centering
	\includegraphics[width=1\linewidth]{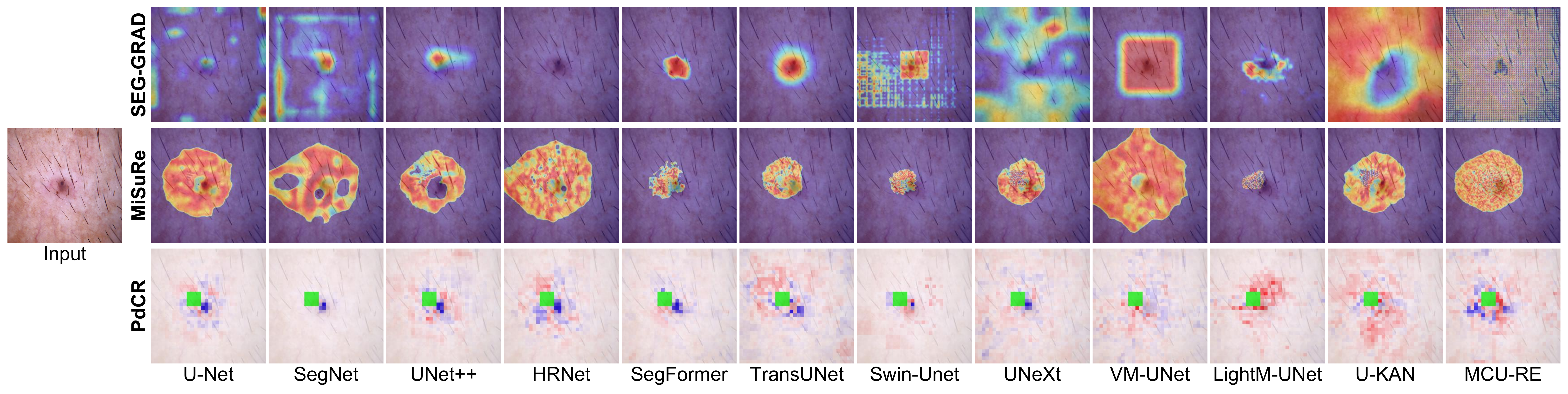}
	\caption{Visualization comparison using three explainability methods. In PdCR, green outlines denote the selected RoI.}
	\label{cmp_exp}
\end{figure*}

Details of the training setup are provided in the Appendix. For causal inference, unless otherwise specified, the initial ITE is computed with $S=3$, and the final ATE is estimated with $N=50$. The RoI size is set to $32 \times 32$, and both the perturbation patch and cropping block size are $8 \times 8$. These hyperparameters are selected based on extensive ablation studies, which are also included in the Appendix.

\begin{table}[t!]
	\centering
	\small
		\begin{tabular}{l|ccc}
			\toprule
			
			\textbf{Method} & \textbf{SEG-GRAD} & \textbf{MiSuRe} & \textbf{PdCR} \\
			\midrule
			\rowcolor[gray]{0.9}
			\multicolumn{4}{c}{HAM10000} \\
			U-Net         & 0.0456 & 0.2115 & \textbf{0.5311} \\
			SegNet        & 0.0929 & 0.1546 & \textbf{0.4047} \\
			UNet++        & 0.1056 & 0.1164 & \textbf{0.3009} \\
			HRNet         & 0.0640 & 0.1424 & \textbf{0.3637} \\
			SegFormer     & 0.0726 & 0.1271 & \textbf{0.2447} \\
			TransUNet     & 0.1726 & 0.1817 & \textbf{0.4181} \\
			Swin-Unet     & 0.0639 & 0.1536 & \textbf{0.3956} \\
			UNeXt         & 0.0646 & 0.2161 & \textbf{0.5990} \\
			VM-UNet       & 0.0776 & 0.0676 & \textbf{0.2083} \\
			LightM-UNet   & 0.1376 & 0.1265 & \textbf{0.2655} \\
			U-KAN         & 0.0433 & 0.1892 & \textbf{0.3443} \\
			MCU-RE        & 0.0907 & 0.1514 & \textbf{0.4048} \\
			\textit{avg.} & 0.0859 & 0.1532 & \textbf{0.3734} \\
			\midrule
			\rowcolor[gray]{0.9}
			\multicolumn{4}{c}{FIVES} \\
			U-Net         & 0.0203 & 0.8615 & \textbf{1.0742} \\
			SegNet        & 0.0357 & 0.2216 & \textbf{0.5027} \\
			UNet++        & 0.0265 & 0.9286 & \textbf{0.9957} \\
			HRNet         & 0.0638 & 0.3512 & \textbf{0.8791} \\
			SegFormer     & 0.0247 & 0.0523 & \textbf{0.2240} \\
			TransUNet     & 0.0902 & 0.2059 & \textbf{0.7746} \\
			Swin-Unet     & 0.0005 & 0.0346 & \textbf{0.0754} \\
			UNeXt         & 0.0816 & 0.4684 & \textbf{0.9554} \\
			VM-UNet       & 0.0113 & 0.0792 & \textbf{0.2384} \\
			LightM-UNet   & 0.0672 & 0.0657 & \textbf{0.3110} \\
			U-KAN         & 0.0626 & 0.4564 & \textbf{0.6228} \\
			MCU-RE        & 0.0621 & 0.7041 & \textbf{0.7420} \\
			\textit{avg.} & 0.0455 & 0.3691 & \textbf{0.6163} \\
			\bottomrule
		\end{tabular}
	\caption{Quantitative evaluation of three explainability methods via averaged attribution analysis on the top-10 salient regions across multiple models (\textit{avg.} denotes the average value).}
	\label{metrics_merged}
\end{table}

\begin{figure}
	\centering
	\includegraphics[width=1\linewidth]{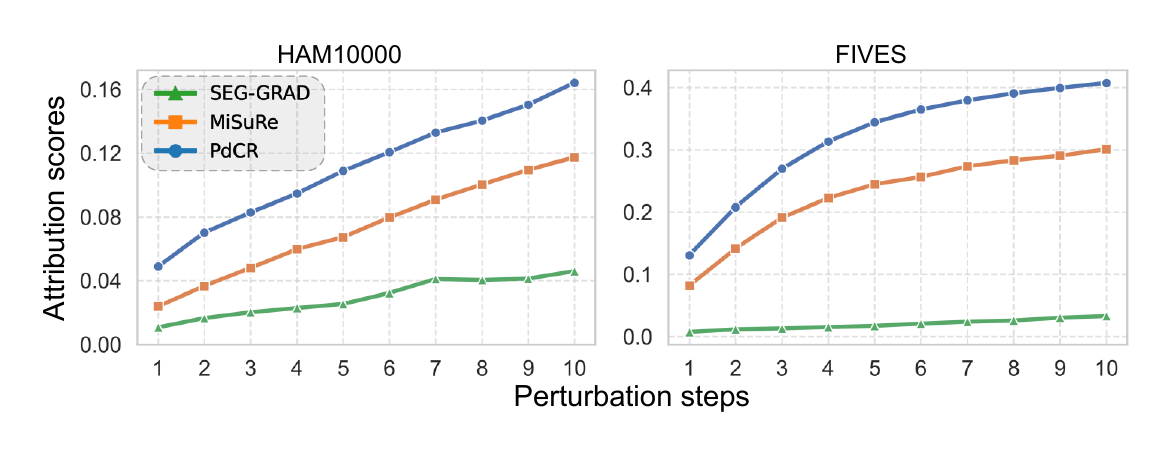}
	\caption{Average attribution curves across datasets under progressive perturbation based on saliency maps.}
	\label{figcmplist}
\end{figure}

\begin{figure*}[t]
	\centering
	\includegraphics[width=1\linewidth]{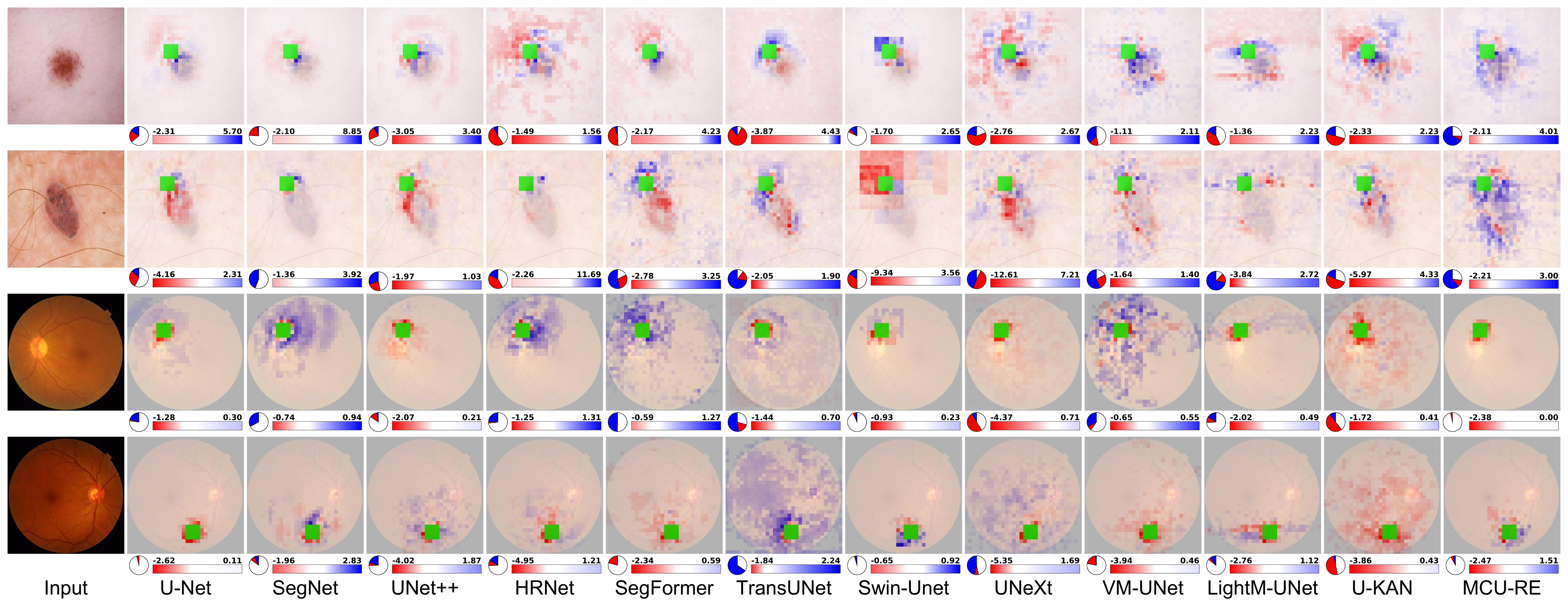}
	\caption{Visualization of PdCR saliency maps for each model on the same RoI. The pie and bar charts below show the proportion of positive (red), negative (blue), and irrelevant (white) regions, with the numeric value indicating the maximum fluctuation in DSC ATE.}
	\label{imgcompare}
\end{figure*}

\subsection{Qualitative Visual Comparison}

First, we visualize the results of three interpretability methods across different models, as shown in Fig.~\ref{cmp_exp}. The saliency maps generated by PdCR exhibit greater consistency in perceiving model mechanisms. 

SEG-GRAD is a white-box explanation method that requires gradient computation at the RoI-corresponding positions of a specific layer (we select the theoretical bottleneck layer of each network). However, the high-level semantic features in segmentation models are often diffuse, and some relevant features may be captured in lower layers. These features do not necessarily retain sufficient spatial detail or provide gradient responses at the RoI (as in the case of HRNet). MiSuRe assumes that the model’s decision for a target is solely determined by surrounding pixels, and hence generates saliency maps by dilating the RoI followed by optimization. This assumption is invalid for models with long-range dependencies. The results of PdCR are obtained through causal reasoning based on input and output, allowing examination of the influence of any position on the RoI.

The comparison methods partially reflect the model’s initial receptive field or inherent characteristics, such as the sliding window in Swin-Unet or the sequential scanning in VM-UNet. However, they fail to reveal deeper insights, nor can they distinguish whether these pixels contribute positively to the RoI prediction. In contrast, PdCR provides a more fine-grained depiction of various deeper model capabilities, which we will further explore in subsequent sections. More qualitative results can be found in the Appendix.

\subsection{Quantitative Attribution Comparison}

\subsubsection{Attribution Scores}
We conducted a quantitative analysis of the attribution metrics~\cite{b0096,b0097} for the baseline methods and PdCR. The numerical results of each model are shown in Table~\ref{metrics_merged}, higher values indicate better performance. Due to its diffuse semantic representations in deeper layers, SEG-GRAD fails to accurately attribute effects in the low-dimensional projection space, resulting in the poorest overall performance. MiSuRe occasionally achieves competitive scores. PdCR consistently outperforms the others in terms of attribution accuracy in the vast majority of cases. This validates the superiority of PdCR in capturing meaningful and fine-grained model attributions, as it can reliably identify the most influential contextual portions for the RoI.

\subsubsection{Attribution Efficiency Trends}
Moreover, the attribution efficiency is analyzed, and the average attribution curves of each method across all models are plotted, as shown in Fig.~\ref{figcmplist}. It can be observed that our method exhibits the fastest rise, followed closely by MiSuRe. In contrast, SEG-GRAD shows the smallest increase, as it cannot provide accurate descriptions of the RoI across all models and may even fail to produce results in some cases.

\begin{table}[t]
	\centering
		\small
		\setlength{\tabcolsep}{1mm} 
		\begin{tabular}{l|cccc}
			\toprule
			\textbf{Method} & \textbf{Pos. (\%)} & \textbf{Neg. (\%)} & \textbf{Irr. (\%)} & 
			\textbf{Min/Max} (\textit{Avg.}) \\
			\midrule
			\rowcolor[gray]{0.9}
			\multicolumn{5}{c}{HAM10000} \\
			
			U-Net        & 14.54 & 14.77 & 70.69 & -4.17 / 3.31 \\
			SegNet       & 17.27 & 19.26 & 63.47 & -3.28 / 2.91 \\
			UNet++       & 19.39 & 24.38 & 56.23 & -2.44 / 3.00 \\
			HRNet        & 24.53 & 25.31 & 50.16 & -3.50 / 3.13 \\
			SegFormer    & 26.14 & 29.93 & 43.93 & -2.84 / 2.41 \\
			TransUNet    & 27.87 & 36.70 & 35.43 & -3.64 / 3.85 \\
			Swin-Unet    & 14.05 & 11.89 & 74.06 & -5.10 / 2.06 \\
			UNeXt        & 35.95 & 32.55 & 31.50 & -3.91 / 5.67 \\
			VM-UNet      & 30.52 & 29.02 & 40.46 & -2.94 / 2.11 \\
			LightM-UNet  & 29.19 & 30.80 & 40.01 & -2.57 / 2.47 \\
			U-KAN        & 30.23 & 24.83 & 40.94 & -3.50 / 2.05 \\
			MCU-RE       & 35.48 & 29.23 & 35.29 & -3.38 / 4.22 \\
			
			\midrule
			\rowcolor[gray]{0.9}
			\multicolumn{5}{c}{FIVES} \\
			
			U-Net        & 8.45  & 7.67  & 83.88 & -5.87 / 0.95 \\
			SegNet       & 13.73 & 16.04 & 70.23 & -5.63 / 2.00 \\
			UNet++       & 16.20 & 11.60 & 72.20 & -8.42 / 2.53 \\
			HRNet        & 15.36 & 18.36 & 66.28 & -6.56 / 2.82 \\
			SegFormer    & 20.83 & 25.74 & 53.43 & -3.51 / 2.06 \\
			TransUNet    & 30.57 & 34.13 & 35.30 & -5.64 / 2.61 \\
			Swin-Unet    & 3.00  & 2.25  & 94.75 & -1.22 / 0.61 \\
			UNeXt        & 37.10 & 29.32 & 33.58 & -9.28 / 3.41 \\
			VM-UNet      & 28.81 & 25.70 & 45.49 & -4.49 / 1.74 \\
			LightM-UNet  & 24.50 & 21.90 & 53.60 & -4.71 / 2.41 \\
			U-KAN        & 40.99 & 28.66 & 30.35 & -5.68 / 2.21 \\
			MCU-RE       & 7.26  & 5.40  & 87.34 & -5.52 / 1.50 \\
			\bottomrule
		\end{tabular}
	\caption{Quantitative comparison of average PdCR across different methods on HAM10000 and FIVES datasets.}
	\label{method_comparison}
\end{table}

\subsection{Model Perception and Reasoning Patterns}

\subsubsection{Model Mechanism Attribution}

We investigate how different segmentation architectures utilize spatial context, revealing that local, global, and sequential mechanisms exhibit distinct patterns of contextual awareness. Fig.~\ref{imgcompare} highlights representative behaviors. CNN-based models largely depend on local neighborhoods, often incorporating nearby pixels. In contrast, architectures with global operations attend to broader pixel contexts. Model-specific observations further illustrate distinctive mechanisms. Compared to U-Net, U-Net++ captures finer details and broader spatial dependencies. Swin-Unet reveals block-like saliency due to its windowed attention, whereas SegFormer and TransUNet exhibit more continuous, global fusion from ViT-style attention. Mamba-based models, such as VM-UNet and MCU-RE, show sequential scanning patterns; VM-UNet links row/column dependencies near RoIs, while LightM-UNet emphasizes horizontal traversal. 

Furthermore, one might expect that the vast majority of patches have non-negative contributions, while our PdCR results suggest otherwise. In Fig.~\ref{imgcompare} and Table~\ref{method_comparison}, the portion of negatively influencing patches, while many reduce the local DSC, is comparable in magnitude to the positive ones. This suggests that deceptive patterns are more pervasive in segmentation problems than commonly recognized, and it reflects the complexity of the underlying segmentation mechanisms. However, a negative influence does not necessarily imply that a patch harms the overall result, as the negative patch for one RoI may also support positively on other RoIs. Verification of whether the patches identified in the PdCR maps truly influence the model’s segmentation output, as well as the application of PdCR to trace decision sources in missegmented regions, is provided in the Appendix.

\subsubsection{Differences in Perception Strategy}

\begin{figure}[t!]
	\centering
	\includegraphics[width=1\linewidth]{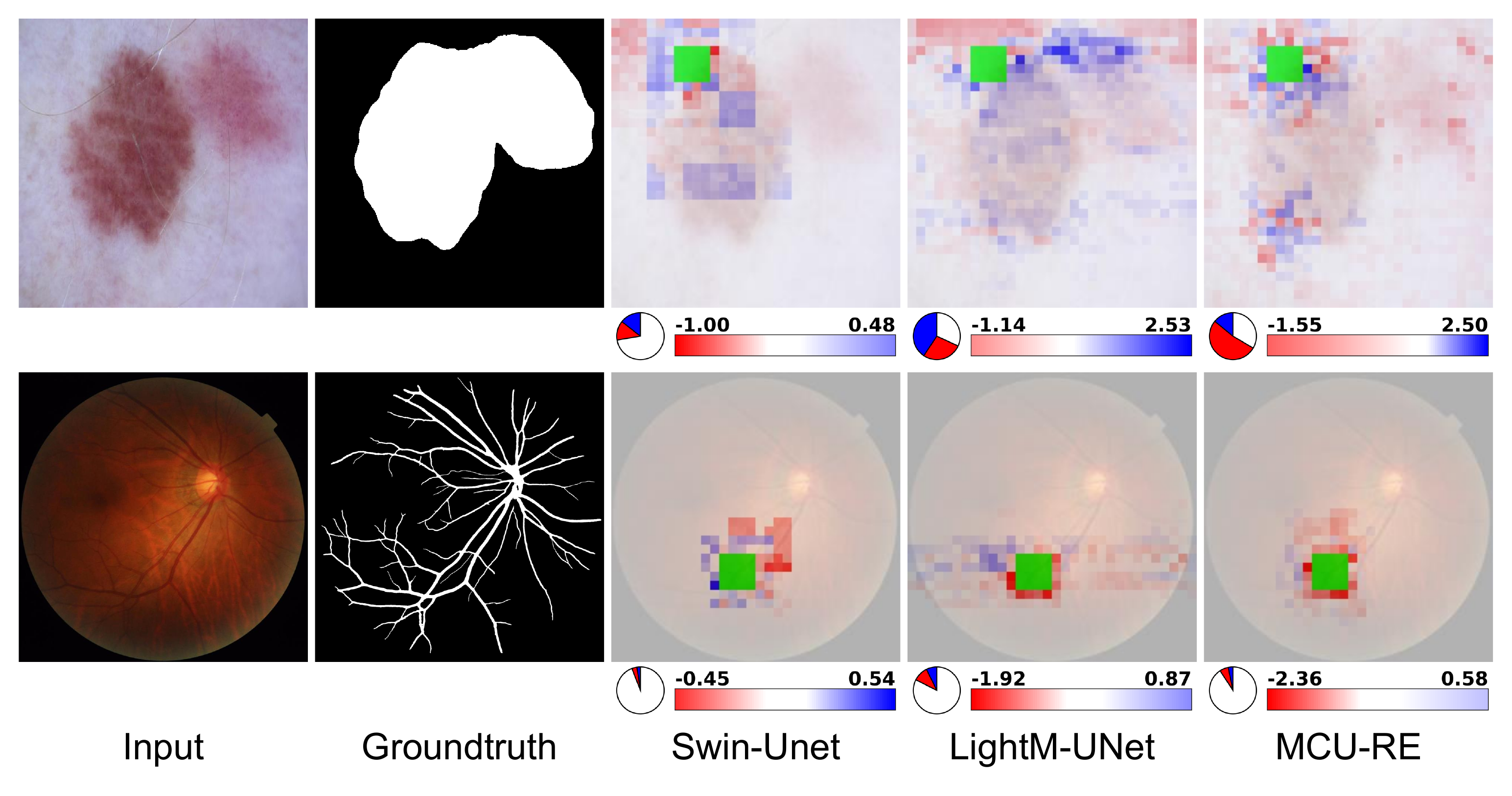}
	\caption{PdCR maps generated by the same model across different datasets reveal markedly different behaviors.}
	\label{mechanismcmp}
\end{figure}

Interestingly, the same algorithm can adopt different perception strategies depending on the dataset, as shown in Fig.~\ref{mechanismcmp}. For example, while segmentation targets in HAM10000 are large and cohesive, those in FIVES are thin, fragmented, and horizontally distributed. Consequently, three  models tend to leverage global context in HAM10000 but shift toward local cues in FIVES. Notably,  MCU-RE, a hybrid of CNN and Mamba, exemplifies this behavior by exhibiting global Mamba-like scanning for skin lesions, while behaving more like a CNN in vessel segmentation, relying on local features and showing a reduced proportion of causally related pixels.

\subsubsection{Quantitative Analysis of Perception Patterns}

To more precisely examine each method’s information-gathering ability, we analyze the spatial context utilization of various segmentation methods via averaged PdCR maps across datasets, revealing distinct patterns of information aggregation tied to architectural design and dataset characteristics.

\begin{figure}
	\centering
	\includegraphics[width=1\linewidth]{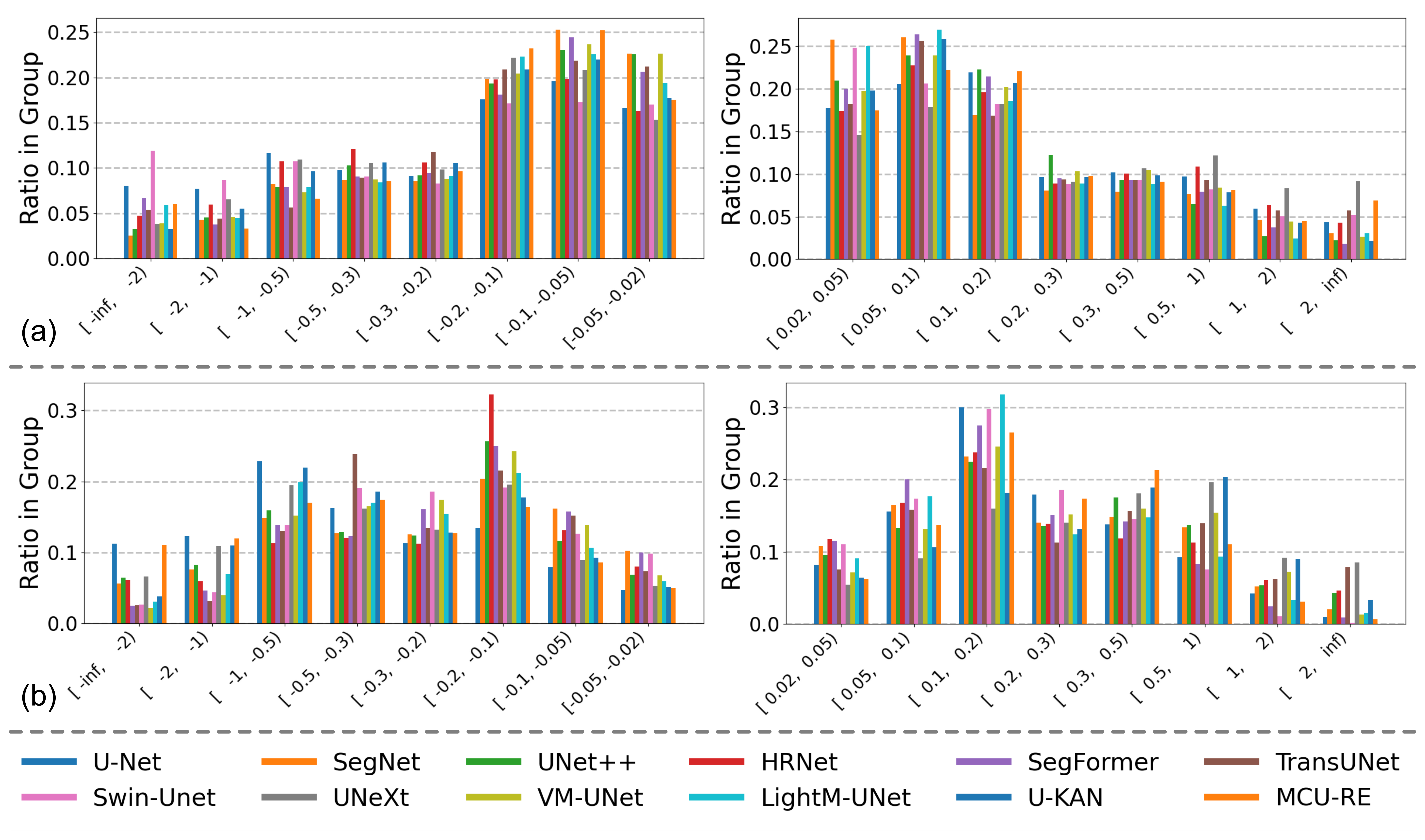}
	\caption{(a) HAM10000 and (b) FIVES datasets: Average proportion of positively and negatively contributing patches across different contribution intervals for each method.}
	\label{percentagecmp}
\end{figure}

Table~\ref{method_comparison} compares the effective receptive fields of models. Early CNNs rely on limited context, which expands with multi-scale features (UNet++, HRNet) or MLPs (UNeXt). Transformers, Mamba, and KAN generally consider larger regions, except Swin-Transformer, which concentrates on a few local windows due to its hierarchical windowing. On HAM10000, positive and negative patch contributions roughly balance, indicating both supportive and suppressive contextual influences. In FIVES, however, most methods show a higher proportion of irrelevant patches and larger absolute decreases, reflecting the challenge of segmenting elongated, fragmented structures. Fig.~\ref{percentagecmp} illustrates the distribution of positive and negative contributions. For connected lesions in HAM10000, perturbations cluster mainly in $(-0.2, -0.02)$ and $(0.02, 0.2)$. In contrast, FIVES shows a broader and more even spread, reflecting the complex, diffuse contextual effects in fine-structure segmentation.

\section{Conclusion}

In this work, we propose PdCR, a model-agnostic method based on causal reasoning to explain medical image segmentation models. Compared with two existing interpretability methods, PdCR achieves superior explanation performance. Through systematic interventions on diverse architectures and datasets, underlying causal relationships and dynamic patterns in model behavior are revealed. These findings highlight the limitations of relying solely on local performance correlations, and underscore the value of causal analysis for interpretability. We hope PdCR will encourage further research into model reliability and causal understanding, moving segmentation models toward greater transparency. Looking forward, integrating causal reasoning with global feature analysis holds promise for developing more trustworthy explainability frameworks.

{
    \small
    \bibliographystyle{unsrt}
    \bibliography{main}
}


\end{document}